\documentclass[a4paper, 10 pt, conference]{ieeeconf}

\IEEEoverridecommandlockouts   
\usepackage{cite}

\usepackage{graphicx}
\graphicspath{{./}{./figures/}{./images_aist/}}
\DeclareGraphicsExtensions{.pdf,.png,.jpeg,.jpg}

\usepackage{amsmath}
\usepackage{amssymb}
\usepackage{array}
\usepackage{url}
\usepackage{xcolor}

\DeclareRobustCommand{\rev}[1]{#1}

\newenvironment{IEEEkeywords}%
  {\vspace{2mm}\noindent\small\textbf{\textit{Index Terms}}---}%
  {\vspace{2mm}}

\title{\LARGE \bf
Online Material Estimation for Conditioned Diffusion Policy\\
in Shaping Deformable Linear Objects}

\author{Ryunosuke~Yamada$^{1}$,
        Tomohiro~Motoda$^{2}$,
        Yukiyasu~Domae$^{2}$,
        and~Tokuo~Tsuji$^{3}$
\thanks{$^{1}$R. Yamada is with the Graduate School of Natural Science and Technology, Kanazawa University, Kanazawa 920-1192, Japan (e-mail: noboriryuu0623@stu.kanazawa-u.ac.jp).}
\thanks{$^{2}$T. Motoda and Y. Domae are with the National Institute of Advanced Industrial Science and Technology (AIST), Tokyo 135-0064, Japan (e-mail: \{tomohiro.motoda, domae.yukiyasu\}@aist.go.jp).}
\thanks{$^{3}$T. Tsuji is with the Faculty of Frontier Engineering, Institute of Science and Engineering, Kanazawa University, Kanazawa 920-1192, Japan (e-mail: tokuo-tsuji@se.kanazawa-u.ac.jp).}}

\begin{document}

\maketitle
\thispagestyle{empty}
\pagestyle{empty}

\begin{abstract}
Shape control of deformable linear objects (DLOs) is challenging for imitation learning because deformation behavior varies with material properties such as stiffness and elasticity, so a single policy must generate different action sequences for different objects even when the goal shape is identical. We propose a diffusion policy conditioned on material labels that are estimated online during manipulation. A recurrent estimation network predicts the material label of the grasped object from the time series of multi-view images and robot joint states, and the predicted label conditions the diffusion policy at every inference step. We collected 480 real-robot demonstrations covering four DLO materials and three groove-placement tasks, and compared per-material specialist policies, a task-conditioned policy without material labels, a policy conditioned on ground-truth material labels, and the proposed policy. Conditioning on ground-truth material labels improved the average success rate from 45.8\% to 60.0\% over the task-only policy, and the proposed policy reached 60.8\% without any prior material information, matching the policy given ground-truth labels. \rev{A post-hoc analysis shows that the estimator extracts material-related information from the manipulation observations and that the diffusion policy responds to the resulting conditioning signal, while the one pronounced failure case is associated with persistent confusion between two similar materials.}
\end{abstract}

\begin{IEEEkeywords}
Deformable Linear Object Manipulation,
Diffusion Policy,
Online Material Estimation,
Conditional Policy,
Imitation Learning
\end{IEEEkeywords}

 
\section{Introduction}

Deformable linear objects (DLOs) such as cables, tubes, and strings are widely encountered in robotic tasks including wire harness assembly, electric wiring, and surgical procedures. Their shape control remains challenging because multiple manipulation sequences can achieve the same goal shape, and objects with different stiffness, elasticity, and surface friction deform differently under identical actions \cite{caporali2026survey}, which makes material-aware manipulation important.

\rev{Classical approaches identify a physical model of the target object for planning and control \cite{azad2023optimal,tabata2023mass,caporali2024deformable,qi2024adaptive}, but require re-identification whenever the object changes and struggle with contact-rich interaction.}

An alternative is imitation learning, which learns manipulation policies directly from demonstrations. It has been applied to DLO tasks such as cable routing and shape manipulation \cite{nair2017combining,luo2024multistage}, and diffusion policies have become a strong general approach for multimodal visuomotor control \cite{chi2025diffusion}. \rev{These approaches learn a policy for a fixed object, while policies that condition on object properties obtain those properties with the help of a simulator \cite{kuroki2024gendom}. In practical environments where a robot manipulates several types of DLOs, no simulator of the object is available and the identity of the presented object is not necessarily known in advance.}

To address this limitation, we propose a manipulation framework that estimates the material label online during manipulation and uses the estimated label as a conditioning input to a diffusion policy. \rev{A material estimation network built on a long short-term memory (LSTM) layer} predicts the material label from the time series of multi-view images and robot joint states, and the estimated label is provided to the diffusion policy at every inference step.

The contributions of this paper are threefold:
\begin{itemize}
    \item \rev{We propose a material-conditioned manipulation framework that at inference requires no simulator, no reference demonstration, and no dedicated probing action: the label is inferred from the task's own observations and re-estimated throughout execution.}
    \item \rev{We demonstrate,} using 480 real-robot demonstrations over four materials and three groove-placement tasks, that conditioning on online estimated material labels achieves manipulation performance comparable to conditioning on ground-truth material labels, without requiring the material label to be specified before execution.
    \item \rev{We show, through a detailed analysis of the recorded rollouts in Section~\ref{sec:analysis}, that the estimator extracts material-related information from the manipulation observations, that the policy responds to the resulting conditioning signal, and that the one pronounced failure case is associated with persistent material confusion.}
\end{itemize}
 
\section{Related Work}

\subsection{Online Adaptation of DLO Manipulation}

\rev{Model-based approaches represent the DLO explicitly, for instance with Cosserat rods \cite{azad2023optimal} or mass--spring systems \cite{tabata2023mass}, but require identification for each object.} To improve adaptability across different DLOs, several studies estimate object-dependent parameters during manipulation. Caporali \emph{et al.} \cite{caporali2024deformable} estimated latent parameters of a learned deformation model online, while Qi \emph{et al.} \cite{qi2024adaptive} adaptively estimated the relationship between robot motion and DLO deformation for shape servoing. These methods adapt internal models or controllers to the manipulated object, whereas our method adapts an imitation-learned policy through online material estimation.

\subsection{Learning-Based Manipulation of DLOs}

Imitation learning has become a promising approach for DLO manipulation. Early work demonstrated that behavior cloning can learn visuomotor policies for rope manipulation from demonstrations \cite{nair2017combining}. Hierarchical imitation learning has further enabled long-horizon DLO manipulation, such as multi-stage cable routing \cite{luo2024multistage}\rev{. More recently, vision-language models have been used for in-context high-level reasoning over DLO routing, with low-level skills trained by reinforcement learning \cite{li2025routing}.}

\subsection{\rev{Estimating Physical Properties through Interaction}}

\rev{A complementary line of work recovers object properties by acting on them: interactive perception exploits properties that become observable only through interaction \cite{bohg2017interactive}, and such properties have been learned from multi-step dynamic interactions \cite{xu2019densephysnet} and from in-hand tactile exploration in SwingBot \cite{wang2020swingbot}.}

\rev{Closer to our setting, the recovered quantity conditions the policy itself. For instance, GenDOM \cite{kuroki2024gendom} conditions a manipulation policy on deformable object parameters identified at inference time. However, it relies on a simulator for both policy training and identification. Furthermore, methods like SwingBot and GenDOM require a dedicated probing action or a reference demonstration, and both identify the object only once rather than continuously.}

\rev{In contrast, our method performs simulator-free estimation directly from the task's own observations and updates the estimate throughout execution. Without a simulator there is no continuous parameter supervision, so we estimate a discrete label instead --- a deliberate trade-off, and Section~\ref{sec:analysis} shows that the estimator need not be perfectly accurate for the conditioning to pay off.}
 
\section{Problem Setting}

We consider a shape-control task in which a robot manipulator places a DLO along a groove of a predefined target shape.

Multiple types of DLOs with different physical properties may be presented, and information about the presented object is not available to the robot before manipulation. One end of the object is fixed to the environment, while the robot initially grasps the other end. The robot manipulates the grasped object and releases it so that the object settles into the target groove. One task episode spans from the initial grasped state to the completion of placement. The initial shape and placement of the object are assumed to be consistent across all object types.

At time $t$, the observation $o_t$ consists of RGB images captured by an environment camera and a hand camera, together with the robot state, comprising joint angles and the gripper state. The action $a_t$ consists of target joint angles and gripper commands. The task is to learn a policy
\[
\pi(a_t \mid o_t, c),
\]
where $c$ denotes the target groove shape, such that the object is successfully placed along the target groove for all considered object types.
 
\section{Proposed Method}
\begin{figure}[tb]
  \centering
  \includegraphics[width=85mm]{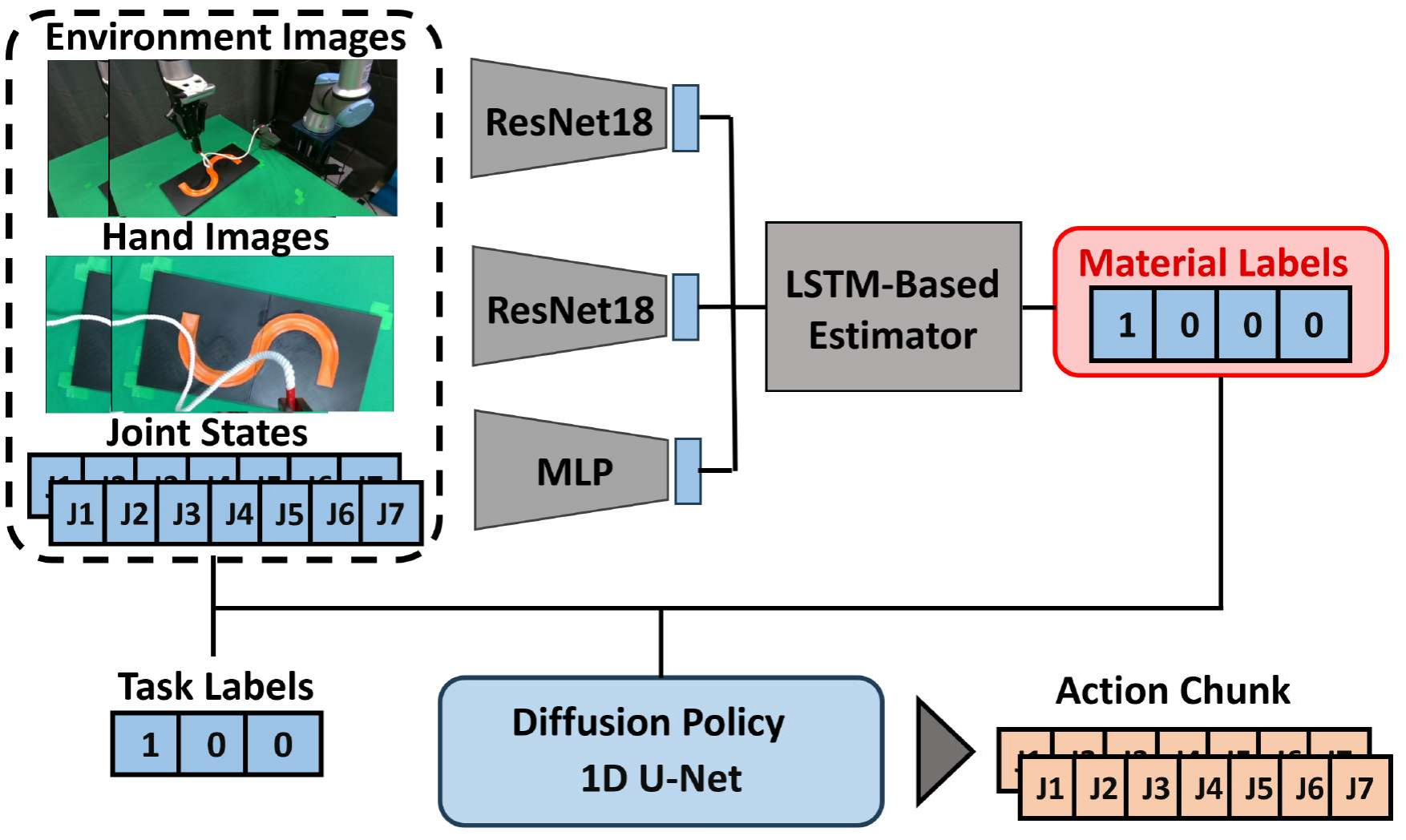}
  \caption{\rev{Overview of the proposed framework. In the upper half, the material estimation network predicts a material
  label from the images and joint states; in the lower half, that label conditions the diffusion policy together with
  the task label. The two networks share no weights, and the label is re-estimated throughout execution.}}
  \label{fig:overview}
\end{figure}
Fig.~\ref{fig:overview} shows the overall pipeline.
The system consists of two networks: a material--task conditioned diffusion policy that generates action sequences, and a material estimation network that predicts the material label online from the interaction history.
\rev{The estimator runs alongside the policy throughout execution, so the conditioning label is refreshed as the manipulation proceeds.}
 
\subsection{Material--Task Conditioned Diffusion Policy}

The deformation behavior of a DLO depends strongly on its material properties, and therefore the appropriate manipulation strategy differs even for the same target shape.
To handle multiple DLO materials with a single policy, we employ a diffusion policy \cite{chi2025diffusion} conditioned on the material and task labels.

RGB images from the environment and hand cameras are encoded by ResNet18 into visual features, which are combined with the robot joint states and one-hot material and task labels\rev{; an observation encoder aggregates these into a global feature used for conditioning}.

The action sequence is generated by iterative denoising with a 1D U-Net conditioned on the global feature.
During training, the ground-truth material label is used for conditioning, whereas during execution it is replaced by the material label estimated by the network described in the next subsection.

\subsection{Online Material Estimation Network}
Materials with different physical properties often exhibit similar visual appearances, making them difficult to distinguish from a single observation. In contrast, their deformation behavior in response to robot manipulation differs depending on the material. The material estimation network therefore predicts the material label from a sequence of observations.

\rev{The latest 10 observations are used as the network input, as shown in the upper half of Fig.~\ref{fig:overview}.} At each time step, RGB images from the environment and hand cameras are encoded by ResNet18, while the robot joint states are encoded by \rev{a multilayer perceptron (MLP)}. The resulting features are concatenated and fed into a single-layer LSTM to capture the temporal interaction history. The LSTM outputs are then aggregated by an attention pooling module, and the resulting feature is mapped by an MLP to the probability of each material label.
The estimator is trained using cross-entropy loss on labeled episodes.

\section{Experiments}

The experiments are conducted to evaluate the effectiveness of the proposed framework from three perspectives.
First, we investigate the material estimation capability by evaluating the influence of training data distributions on the estimation model.
Second, we evaluate the manipulation performance of the material-conditioned diffusion policy by comparing policies trained under different conditioning settings and material information availability.
Finally, \rev{we analyze the recorded rollouts to characterize the online estimator and its effect on the generated actions.}

\subsection{Setup and Data Collection}
\begin{figure}[tb]
  \centering
  \includegraphics[width=70mm]{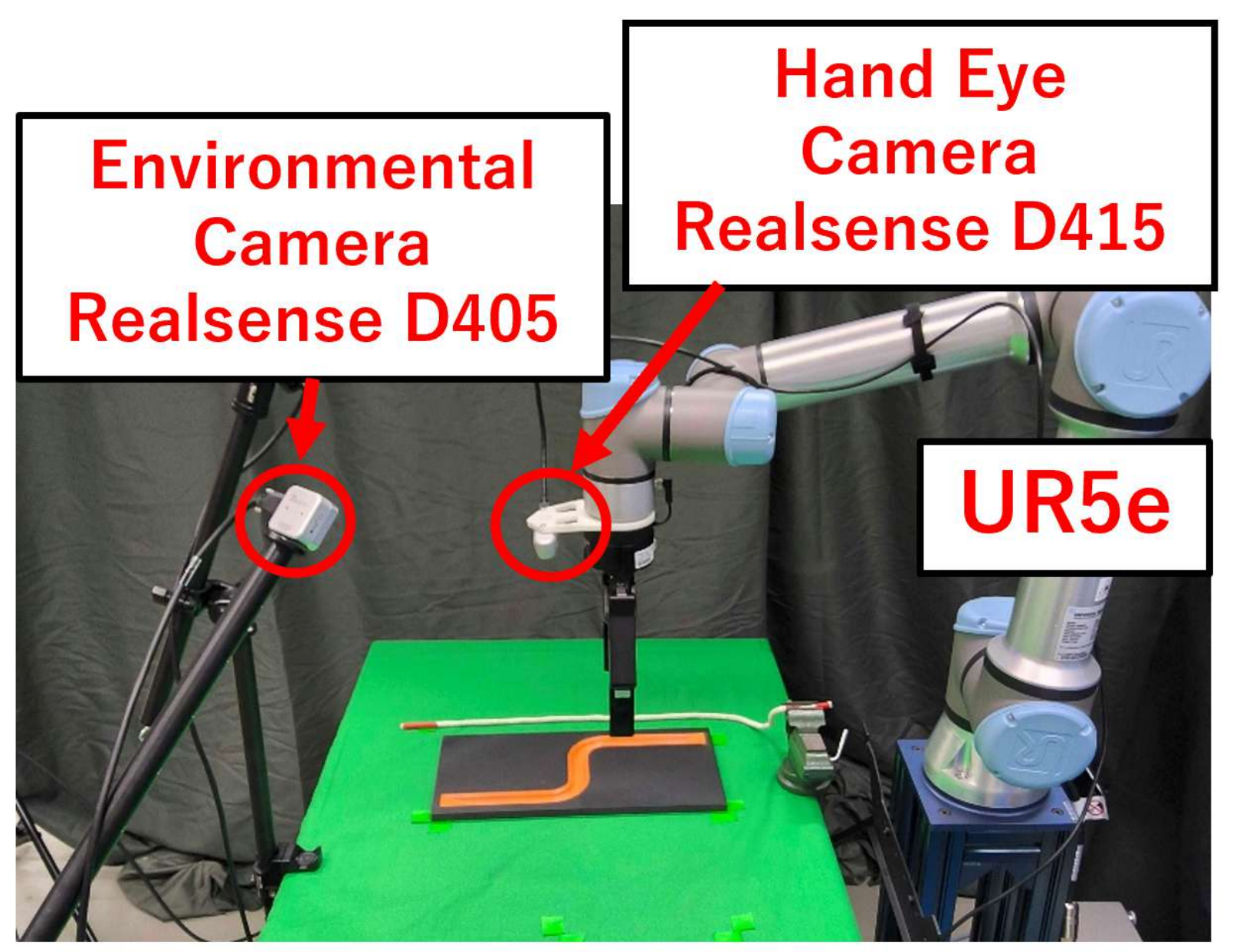}
  \vspace{-2mm}
  \caption{Real-robot experimental environment.}
  \label{fig:env}
\end{figure}
\begin{figure}[tb]
  \centering
  \includegraphics[width=65mm]{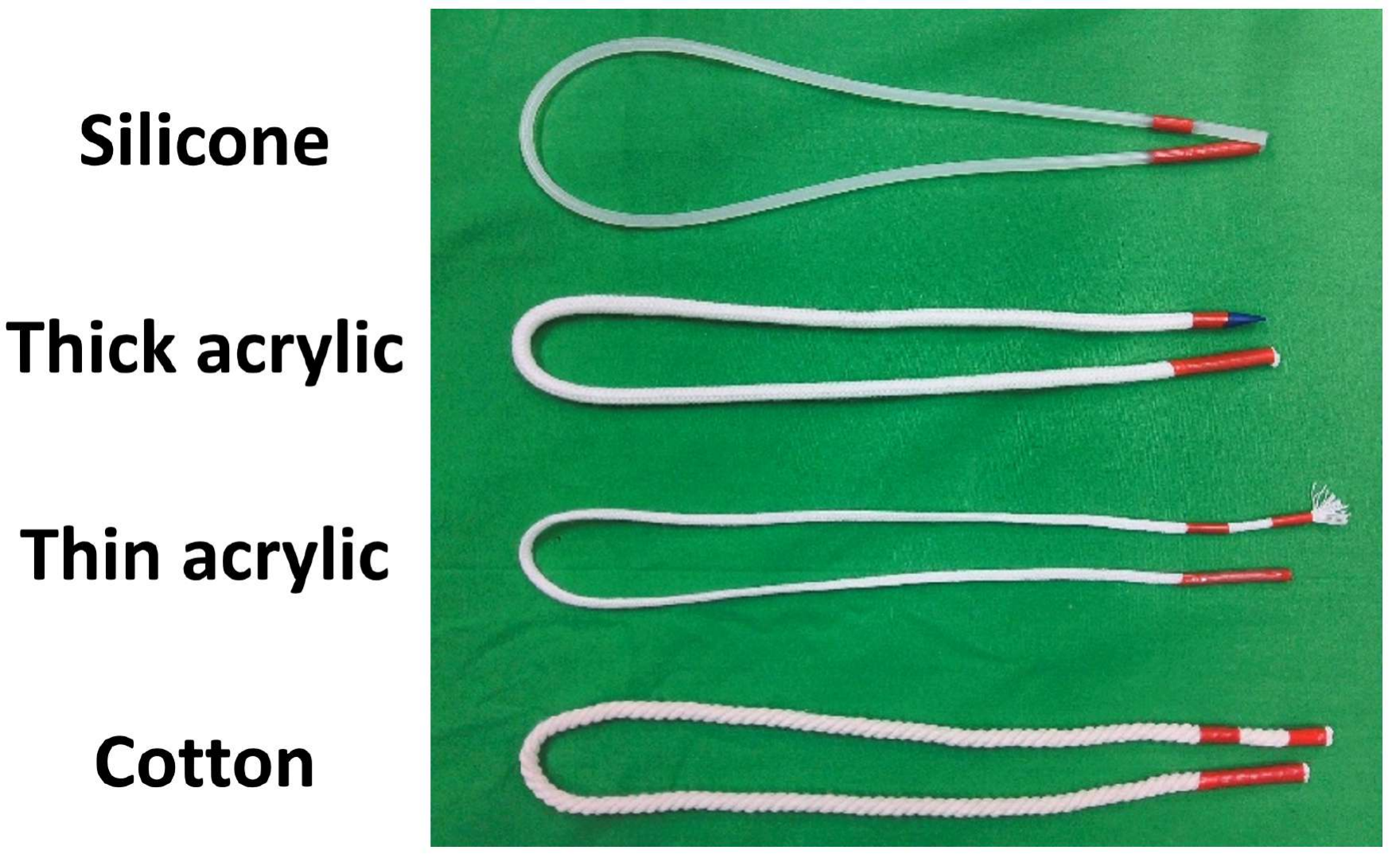}
  \vspace{-2mm}
  \caption{The four DLOs used in the tasks.}
  \label{fig:dlos}
\end{figure}
\begin{figure}[t]
  \centering
  \includegraphics[width=70mm]{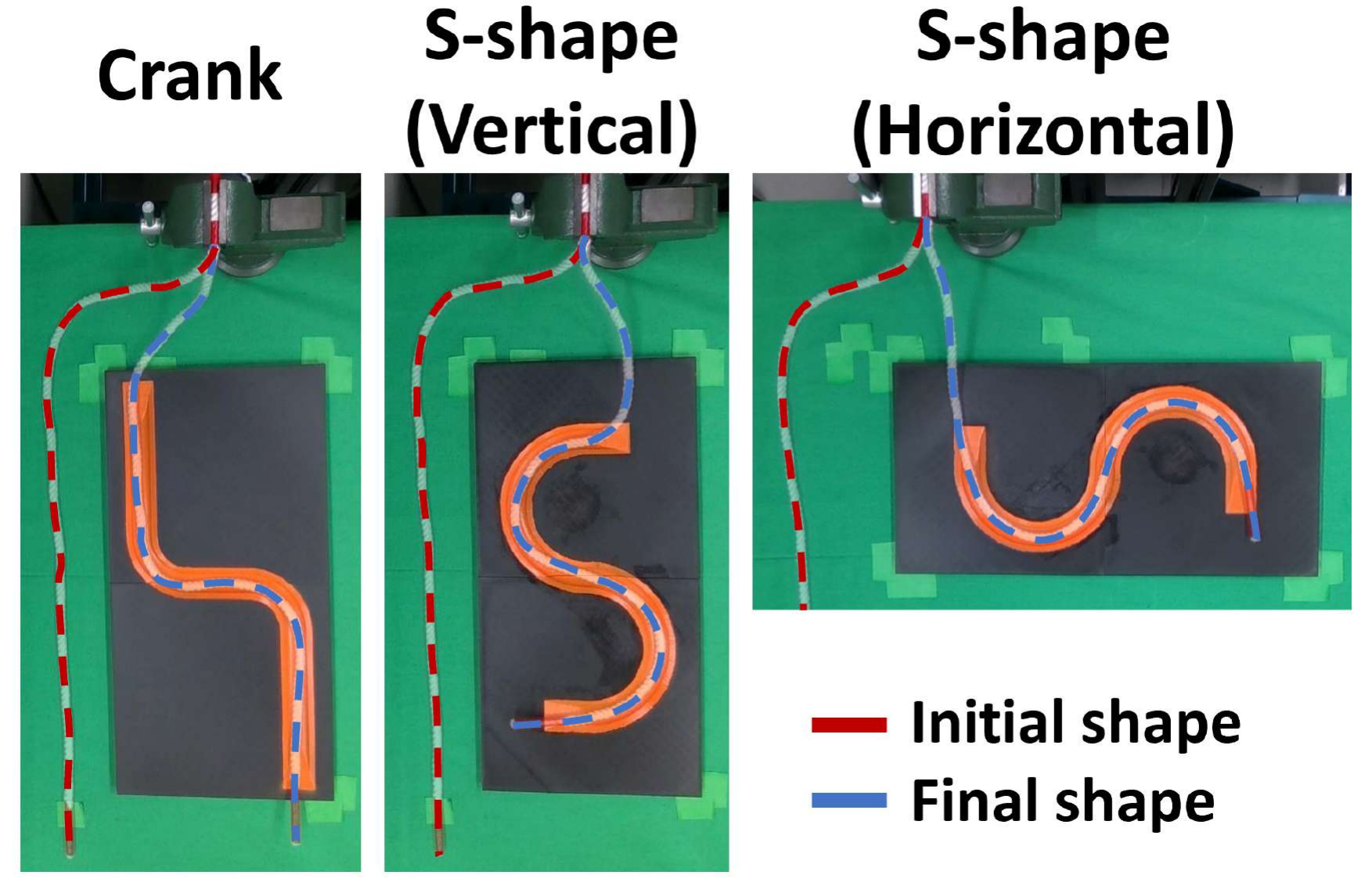}
  \vspace{-2mm}
  \caption{Initial and goal configurations for each task board.}
  \label{fig:tasks}
\end{figure}
Fig.~\ref{fig:env} shows the real-robot experimental environment.
A 6-\rev{degree-of-freedom} UR5e manipulator is used with two RGB cameras: a RealSense D405 mounted as an environment camera and a RealSense D415 mounted near the robot hand.

To evaluate the adaptability of the proposed framework to different physical properties, four DLOs with different materials are prepared: Silicone, Thick acrylic, Thin acrylic, and Cotton, shown in Fig.~\ref{fig:dlos}.
Next, three groove-placement tasks are designed using two task boards with different target shapes: a crank-shaped groove and an S-shaped groove.
The three tasks are denoted C for crank, SV for vertical S, and SH for horizontal S, where SV and SH use the same S-shaped groove board with different initial configurations, as shown in Fig.~\ref{fig:tasks}.

The groove depth is designed to accommodate the thickest DLO.
The groove cross-section consists of a circular bottom and inclined surfaces, which makes the DLO more likely to deviate from the groove when excessive tension is applied.
This design allows the experiments to evaluate whether the learned policies can generate appropriate manipulation strategies according to the material-dependent deformation behavior.

Expert demonstrations are collected through teleoperation using a 3Dconnexion SpaceMouse.
For each of the 12 material--task combinations, 40 successful episodes are collected, resulting in a total of 480 demonstrations.

\subsection{Compared Policies and Evaluation Protocol}

To evaluate the effectiveness of material conditioning and online material estimation, we compare four diffusion policies implemented using RoboManipBaselines \cite{murooka2026robomanip}. The following names are used throughout the paper.

\begin{enumerate}
    \item \textbf{Specialist policy}: one diffusion policy per material--task pair, trained on the corresponding 40 demonstrations.
    \item \textbf{Task-only policy}: a single policy trained on all 480 demonstrations and conditioned only on the task label, isolating the effect of removing material information.
    \item \textbf{Oracle policy}: the same, additionally conditioned on the ground-truth material label, representing the case where the material identity is known in advance.
    \item \textbf{Estimated policy}: the proposed policy. It uses the Oracle weights, with the material label replaced by the online estimate at inference.
\end{enumerate}
 
\rev{Images are cropped from $160 \times 120$ to $128 \times 96$ pixels. The policy predicts 16 actions from the past two observations and executes the first eight, sampled with an 8-step denoising diffusion implicit model schedule; training used a batch size of 256, a learning rate of $5 \times 10^{-5}$, and 250 epochs, and the final-epoch checkpoint was evaluated.}

A trial is classified as \emph{success} when \rev{the released DLO lies within the outline of the groove as seen from above, regardless of its height above the board, and as a \emph{failure} otherwise}.

Evaluation rollouts are conducted for each policy with 10 trials for each of the 12 material--task combinations, resulting in 120 trials per policy.
The initial DLO configurations and fixed-end poses are approximately matched across policies, and the random seeds are kept identical.
 
\subsection{Evaluation of the Material Estimation Network}

Before evaluating the manipulation performance, we evaluate the material estimation network on a held-out dataset of 120 real-robot rollouts generated by the Oracle policy covering four materials and three tasks.
The estimator receives the latest 10 observations and predicts the material label of the manipulated DLO.
The classification accuracy is calculated by comparing the predicted material label with the ground-truth material label for each observation sample.

The observation distribution used to train the estimator affects its ability to recognize material-dependent deformation behavior.
We compared three types of training datasets:

\begin{itemize}
    \item \textbf{Teleoperation}: human-operated demonstrations,
    \item \textbf{Task-only rollout}: real-robot rollouts generated by a diffusion policy conditioned only on the task label,
    \item \textbf{Specialist rollout}: real-robot rollouts generated by Specialist policies trained separately for each material--task pair.
\end{itemize}

All estimators were trained with the same network architecture and evaluated on the Oracle policy rollouts described above.
Table~\ref{tab:estimator_dataset} summarizes the results.

\begin{table}[t]
\caption{Material estimation accuracy for different training datasets.}
\begin{center}
\begin{tabular}{|l|c|}
\hline
\textbf{Training dataset} & \textbf{Accuracy [\%]} \\
\hline
Teleoperation & 73.34 \\
\hline
Task-only rollout & 66.18 \\
\hline
Specialist rollout & \textbf{79.28} \\
\hline
\end{tabular}
\label{tab:estimator_dataset}
\end{center}
\end{table}

The Specialist rollout dataset achieved the highest accuracy, outperforming the other datasets by more than 5 percentage points.
This indicates that trajectories generated by policies specialized for each material--task pair contain clearer material-dependent deformation characteristics than human demonstrations or the Task-only policy.
Therefore, Specialist rollout data was used for training the final estimator.

\rev{Architectural variants were also compared: removing attention pooling, raising dropout from 0.15 to 0.25, and reducing the encoder and LSTM feature dimensions. The smallest model generalized best, achieving an estimation accuracy of 96.26\%, and was used throughout.}
 
\subsection{Manipulation Results}

\begin{table}[t]
\caption{Average success rate [\%] per material for the Specialist, Task-only, Oracle, and Estimated policies.}
\begin{center}
\begin{tabular}{|l|c|c|c|c|}
\hline
\textbf{Material} & \textbf{Specialist} & \textbf{Task-only} & \textbf{Oracle} & \textbf{Estimated} \\
\hline
Silicone & 56.7 & 56.7 & \textbf{73.3} & 40.0 \\
\hline
Thick acrylic & 33.3 & 23.3 & 46.7 & \textbf{56.7} \\
\hline
Thin acrylic & 50.0 & 63.3 & 53.3 & \textbf{70.0} \\
\hline
Cotton & 26.7 & 40.0 & 66.7 & \textbf{76.7} \\
\hline
Average & 41.7 & 45.8 & 60.0 & \textbf{60.8} \\
\hline
\end{tabular}
\label{tab:results}
\end{center}
\end{table}

\begin{figure*}[t]
  \centering
  \includegraphics[width=0.94\textwidth]{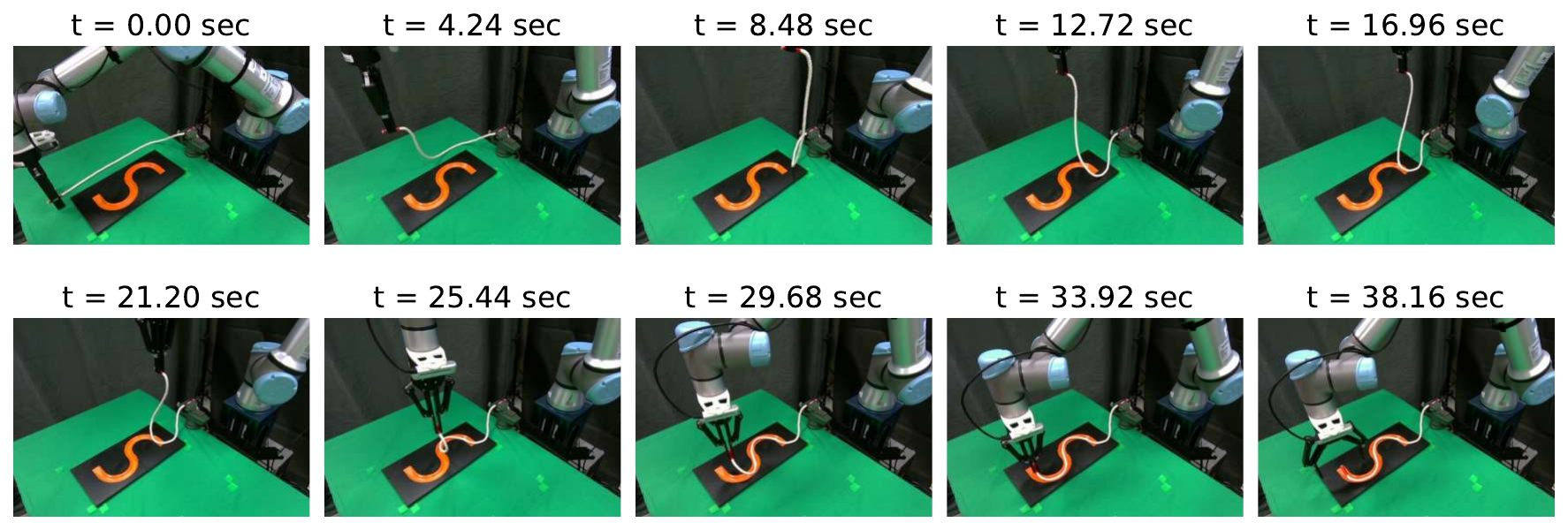}
  \vspace{-2mm}
  \caption{Manipulation example by the Estimated policy on the SV task with Cotton.}
  \label{fig:timelapse}
\end{figure*}

\begin{figure}[t]
  \centering
  \includegraphics[width=85mm]{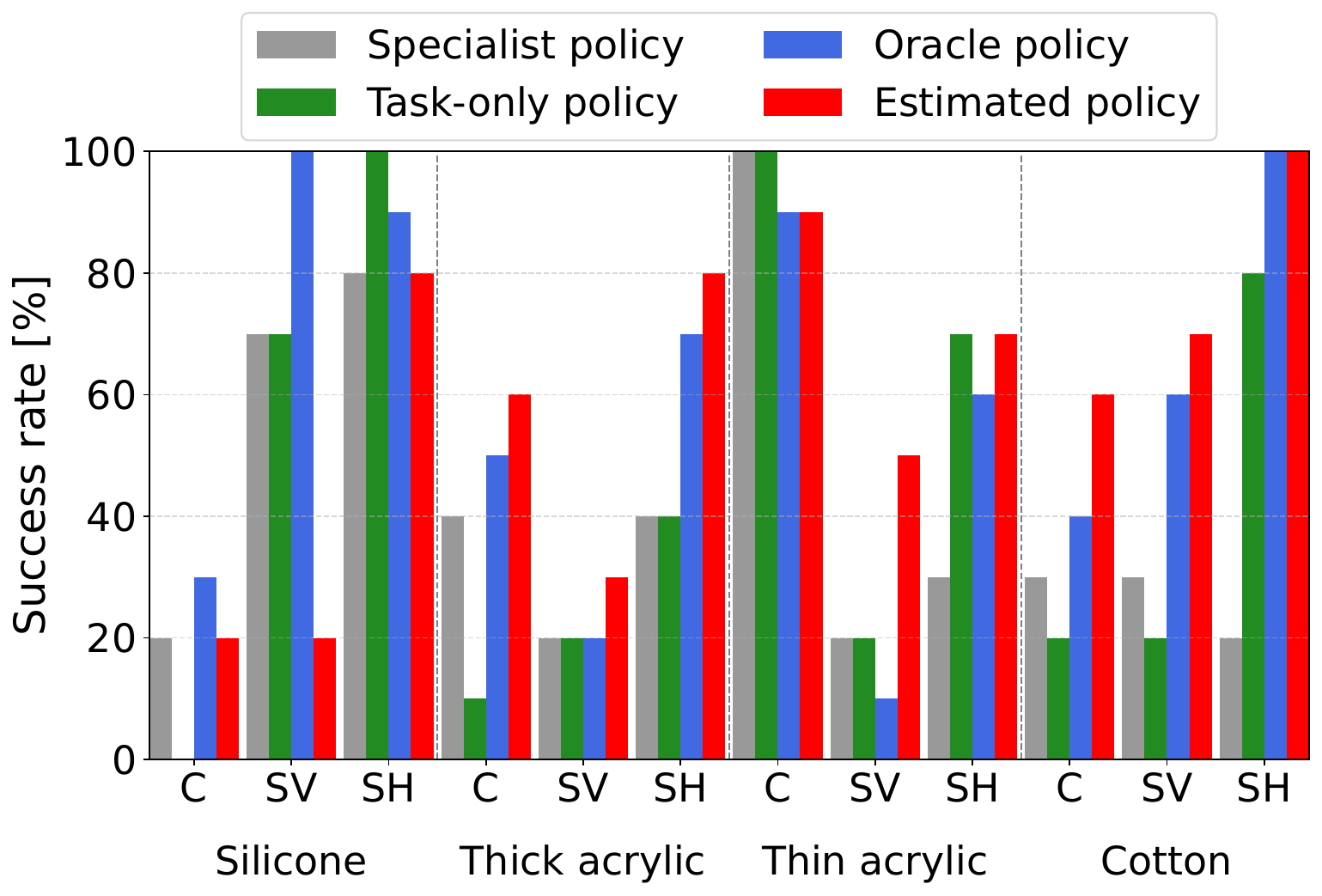}
  \vspace{-2mm}
  \caption{\rev{Success rate of the four policies for each material and task.}}
  \label{fig:estimated_comparison}
\end{figure}

Fig.~\ref{fig:timelapse} shows an example of successful manipulation performed by the Estimated policy.
Table~\ref{tab:results} summarizes the average success rates of the four policies over all material--task combinations, and \rev{Fig.~\ref{fig:estimated_comparison} breaks them down by material and task.}
Overall, the proposed Estimated policy achieved an average success rate of 60.8\%, which is comparable to the Oracle policy at 60.0\%, despite the absence of prior material information.

Both the Oracle and Estimated policies significantly outperformed the Specialist policy at 41.7\% and the Task-only policy at 45.8\%.
This advantage over the Specialist policy is likely because the unified models leverage all 480 demonstrations across materials while retaining material-specific conditioning, whereas the Specialist policy is trained on only 40 demonstrations per pair.
Compared with the Task-only policy, providing material conditioning, whether ground-truth or estimated, improved performance, particularly for Silicone and Cotton where material-dependent strategies are critical.

However, the per-material results show variation across materials.
For Thick acrylic, Thin acrylic, and Cotton, the Estimated policy outperformed the Oracle policy by \rev{10 to 17} percentage points, whereas for Silicone it achieved 40.0\% compared to 73.3\% of the Oracle policy.
 
\subsection{Analysis of Estimator and Policy Behavior}
\label{sec:analysis}

\rev{The Estimated policy reaches the same average success rate as the Oracle policy, yet the label it feeds to the
policy is often wrong. We therefore analyze the 120 rollouts recorded for the Estimated policy.}

\begin{figure}[tb]
  \centering
  \includegraphics[width=85mm]{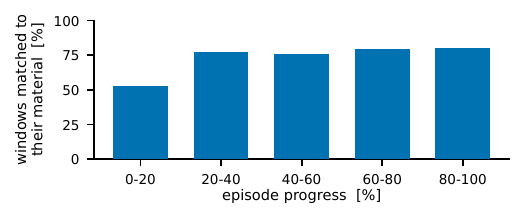}
  \vspace{-2mm}
  \caption{\rev{Fraction of windows that $k$-means groups with their own material, for each fifth of the episode.
  The clustering is given no access to the material label.}}
  \label{fig:clustering}
\end{figure}

\rev{We first evaluate whether the manipulation observations carry material information, separately from whether
the estimator names the material correctly. The estimator's own output does not answer this question: it is
correct for $67\%$ of the windows, and this value varies by less than two percentage points over the episode
because its errors are systematic rather than disorganized. They fall into two pairs, Silicone with Thin acrylic
and Thick acrylic with Cotton, and rarely cross between the pairs, so an incorrect label does not imply that the
material is poorly separated in the representation. We therefore evaluate the arrangement of the representation
itself, using the vector that the estimator pools from its per-step features before the final classification layer.
Each episode is divided into five equal intervals, and within each interval the windows are grouped into four
clusters by $k$-means, without access to the material label. The four clusters are then assigned to the four
materials by the one-to-one assignment that agrees best with the ground truth, and we report the fraction of
windows that this assignment places with their own material; an arbitrary grouping would place $25\%$. As shown in
Fig.~\ref{fig:clustering}, this fraction is $52.9\%$ in the first fifth of the episode and $80.2\%$ in the last
fifth. The material is therefore recoverable from the representation without supervision, and increasingly so as
the manipulation proceeds, which is consistent with material properties becoming observable through deformation
under contact rather than from appearance alone.}

\begin{figure}[tb]
  \centering
  \includegraphics[width=85mm]{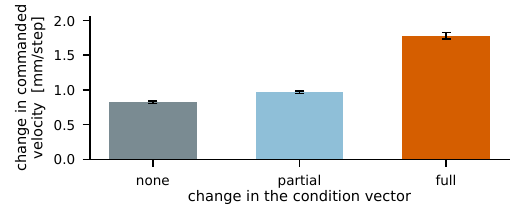}
  \vspace{-2mm}
  \caption{\rev{Change in commanded end-effector velocity across an action-chunk boundary, grouped by how far the
  condition vector moved across that boundary: not at all, without changing which material has the highest
  probability, and enough to change it. Bars are means over the 7448 boundaries, with standard errors.}}
  \label{fig:response}
\end{figure}

\rev{We next evaluate whether the policy acts on the estimate it is given. The Task-only policy receives the same
images but does not reach the same success rate, so what distinguishes the proposed policy is the explicit
conditioning input: a four-dimensional vector of material probabilities, recomputed every 24 steps when the
previous action chunk has been executed. Across such a boundary the scene, the robot, and the object change only
gradually, so an abrupt change in the condition vector is a plausible cause of an abrupt change in the generated
action. The 7448 boundaries are grouped by how far the condition vector moved, and the change in commanded
end-effector velocity is measured across each of them, as shown in Fig.~\ref{fig:response}. The velocity changed
by $0.82$ mm/step where the condition vector did not move, by $0.97$ mm/step where it moved without changing which
material had the highest probability, and by $1.78$ mm/step where that material changed, corresponding to factors
of $1.17$ and $2.17$ relative to no movement. The policy therefore responds not only when the predicted material
changes, but also to smaller movements that leave the prediction unchanged.}

\rev{Finally, we examine the one combination in which the Estimated policy did not match the Oracle policy.
As shown in Fig.~\ref{fig:estimated_comparison}, Silicone on the SV task is the only combination in which it fell
clearly below the others, achieving $20\%$ against $70\%$ for both the Specialist and the Task-only policy and
$100\%$ for the Oracle policy. Since two policies that receive no estimated label both achieve $60\%$, the
difficulty of this combination alone is unlikely to explain the drop. Over its ten trials, the estimator assigned
Silicone an average probability of $0.81$ in the two trials that succeeded but only $0.32$ in the eight that
failed, assigning $0.66$ to Thin acrylic instead. The number of trials is too small for a statistical test, but
Silicone and Thin acrylic are the pair that the estimator separates least well, and in this combination the
confusion persisted through most of the episode.}

\section{Conclusion}

We proposed a framework for shape control of DLOs in which a material label is estimated online from the interaction history and conditions a diffusion policy in real time.
On a real-robot groove-placement benchmark with four materials and three tasks, the Oracle policy, conditioned on ground-truth material labels, improved the average success rate from 45.8\% to 60.0\% over a task-only ablation, and the proposed Estimated policy, conditioned on online material estimation, achieved a comparable success rate of 60.8\% without any prior material information.
\rev{A post-hoc analysis indicates that the estimator extracts material-related information from the manipulation observations, that the diffusion policy changes its generated action in proportion to how much the conditioning signal moves, and that the one pronounced failure case is associated with persistent confusion between two similar materials.}
Future work includes generalization to unseen materials, and replacing the discrete material label with a higher-dimensional, interpretable material representation whose dimensions correspond to quantities such as friction and elasticity, to enable explainable action generation.
 
\section*{Acknowledgment}
This work was supported by JSPS KAKENHI Grant Number JP26KJ1265.
The authors thank the Embodied AI Research Team, Artificial Intelligence Research Center, AIST, for their support and for providing the experimental environment and computational resources.


\begin{thebibliography}{99}

\bibitem{caporali2026survey}
A. Caporali, I. Cuiral-Zueco, G. López-Nicolás, and G. Palli,
``Robotic perception and manipulation of deformable linear objects: A survey,''
\textit{The International Journal of Robotics Research}, 2026, OnlineFirst.

\bibitem{azad2023optimal}
A. Artinian, Q. Huet, F. Ben Amar, and V. Perdereau,
``Optimal Cosserat-based deformation control for robotic manipulation of linear objects,''
in \textit{Proc. IEEE/ASME Int. Conf. Advanced Intelligent Mechatronics (AIM)},
2023, pp. 381--388.

\bibitem{tabata2023mass}
K. Tabata, H. Seki, T. Tsuji, et al.,
``Mass spring model for non-uniformed deformable linear object toward dexterous manipulation,''
\textit{Artificial Life and Robotics}, vol. 28, pp. 812--822, 2023.

\bibitem{caporali2024deformable}
A. Caporali, P. Kicki, K. Galassi, R. Zanella, K. Walas, and G. Palli,
``Deformable linear objects manipulation with online model parameters estimation,''
\textit{IEEE Robotics and Automation Letters}, vol. 9, no. 3, pp. 2598--2605, 2024.

\bibitem{qi2024adaptive}
J. Qi, G. Ran, B. Wang, J. Liu, W. Ma, P. Zhou, and D. Navarro-Alarcon,
``Adaptive shape servoing of elastic rods using parameterized regression features and auto-tuning motion controls,''
\textit{IEEE Robotics and Automation Letters}, vol. 9, no. 2, pp. 1428--1435, 2024.

\bibitem{nair2017combining}
A. Nair, D. Chen, P. Agrawal, P. Isola, P. Abbeel, J. Malik, and S. Levine,
``Combining self-supervised learning and imitation for vision-based rope manipulation,''
in \textit{Proc. IEEE Int. Conf. Robotics and Automation (ICRA)},
2017, pp. 2146--2153.

\bibitem{luo2024multistage}
J. Luo, C. Xu, X. Geng, G. Feng, K. Fang, L. Tan, S. Schaal, and S. Levine,
``Multistage cable routing through hierarchical imitation learning,''
\textit{IEEE Transactions on Robotics}, vol. 40, pp. 1476--1491, 2024.

\bibitem{chi2025diffusion}
C. Chi, Z. Xu, S. Feng, et al.,
``Diffusion policy: Visuomotor policy learning via action diffusion,''
\textit{The International Journal of Robotics Research}, vol. 44, no. 10--11,
pp. 1684--1704, 2025.

\bibitem{kuroki2024gendom}
S. Kuroki, J. Guo, T. Matsushima, T. Okubo, M. Kobayashi, Y. Ikeda,
R. Takanami, P. Yoo, Y. Matsuo, and Y. Iwasawa,
``GenDOM: Generalizable one-shot deformable object manipulation with parameter-aware policy,''
in \textit{Proc. IEEE Int. Conf. Robotics and Automation (ICRA)},
2024, pp. 14792--14799.

\bibitem{li2025routing}
M. Li, H. Yu, Y. Huang, Y. Hong, H. Ye, and C. Choi,
``Hierarchical DLO routing with reinforcement learning and in-context vision-language models,''
\textit{arXiv preprint arXiv:2510.19268}, 2025.

\bibitem{bohg2017interactive}
J. Bohg, K. Hausman, B. Sankaran, O. Brock, D. Kragic, S. Schaal,
and G. S. Sukhatme,
``Interactive perception: Leveraging action in perception and perception in action,''
\textit{IEEE Transactions on Robotics}, vol. 33, no. 6, pp. 1273--1291, 2017.

\bibitem{xu2019densephysnet}
Z. Xu, J. Wu, A. Zeng, J. B. Tenenbaum, and S. Song,
``DensePhysNet: Learning dense physical object representations via multi-step dynamic interactions,''
in \textit{Proc. Robotics: Science and Systems (RSS)}, 2019.

\bibitem{wang2020swingbot}
C. Wang, S. Wang, B. Romero, F. Veiga, and E. Adelson,
``SwingBot: Learning physical features from in-hand tactile exploration for dynamic swing-up manipulation,''
in \textit{Proc. IEEE/RSJ Int. Conf. Intelligent Robots and Systems (IROS)},
2020, pp. 5633--5640.

\bibitem{murooka2026robomanip}
M. Murooka, T. Motoda, R. Nakajo, H. Oh, K. Makihara, K. Shirai,
T. Ogata, and Y. Domae,
``RoboManipBaselines: A unified framework for imitation learning in robotic manipulation across real and simulation environments,''
\textit{IEEE Access}, vol. 14, pp. 97896--97909, 2026.

\end{thebibliography}
\end{document}